\documentclass[10pt,twocolumn,letterpaper]{article}

\usepackage{cvpr}              
\definecolor{cvprblue}{rgb}{0.21,0.49,0.74}
\usepackage[pagebackref,breaklinks,colorlinks,allcolors=cvprblue]{hyperref}
\usepackage{multirow}
\usepackage{graphicx}
\usepackage{minted}

\def\paperID{39778} 
\def\confName{CVPR}
\def\confYear{2026}

\begin{document}

\title{VLM-Guided Group Preference Alignment for Diffusion-based \\ Human Mesh Recovery}

\author{
    Wenhao Shen$^{1}$ \quad 
    Hao Wang$^{2\dag}$ \quad 
    Wanqi Yin$^{3}$ \quad 
    Fayao Liu$^{4}$ \quad
    Xulei Yang$^{4}$ \\
    Chao Liang$^{1}$ \quad 
    Zhongang Cai$^{3}$ \quad 
    Guosheng Lin$^{1\dag}$ \\
    $^1$Nanyang Technological University \quad 
    $^2$HKUST(GZ) \quad 
    $^3$SenseTime Research \quad
    $^4$A*STAR
}

\maketitle
\let\thefootnote\relax\footnotetext{$^\dag$Corresponding authors.}

\begin{abstract}
Human mesh recovery (HMR) from a single RGB image is inherently ambiguous, as multiple 3D poses can correspond to the same 2D observation. Recent diffusion-based methods tackle this by generating various hypotheses, but often sacrifice accuracy. 
They yield predictions that are either physically implausible or drift from the input image, especially under occlusion or in cluttered, in-the-wild scenes. 
To address this, we introduce a dual-memory augmented HMR critique agent with self-reflection to produce context-aware quality scores for predicted meshes. 
These scores distill fine-grained cues about 3D human motion structure, physical feasibility, and alignment with the input image. We use these scores to build a group-wise HMR preference dataset.
Leveraging this dataset, we propose a group preference alignment framework for finetuning diffusion-based HMR models. This process injects the rich preference signals into the model, guiding it to generate more physically plausible and image-consistent human meshes.
Extensive experiments demonstrate that our method achieves superior performance compared to state-of-the-art approaches.
\end{abstract}
    
\section{Introduction}
\label{sec:intro}

This paper aims to tackle the challenge of monocular human mesh recovery (HMR), which involves estimating both human pose and shape from a single RGB image. Accurate 3D human perception is crucial for understanding activities and enabling applications such as VR, robotics, and gaming. However, this task is inherently ill-posed, as lifting 2D observations to 3D motion is highly ambiguous.

Specifically, optimization-based methods~\cite{bogo2016keepitsmpl, pavlakos2019smplxehf, joo2021eft} iteratively refine human pose and shape from 2D cues, while regression-based approaches~\cite{goel2023hmr2.0, sarandi2024nlf, wang2025prompthmr, patel2025camerahmr} directly predict a single mesh per image. Both struggle with the ambiguity of lifting 2D to 3D, especially under occlusion or depth uncertainty. In contrast, probabilistic methods~\cite{kolotouros2021prohmr, sengupta2023humaniflow, xu2024scorehypo, shen2025adhmr, fiche2025mega} model this uncertainty via multiple predictions but often sacrifice accuracy for diversity.

\begin{figure}
    \centering
    \includegraphics[width=0.95\linewidth]{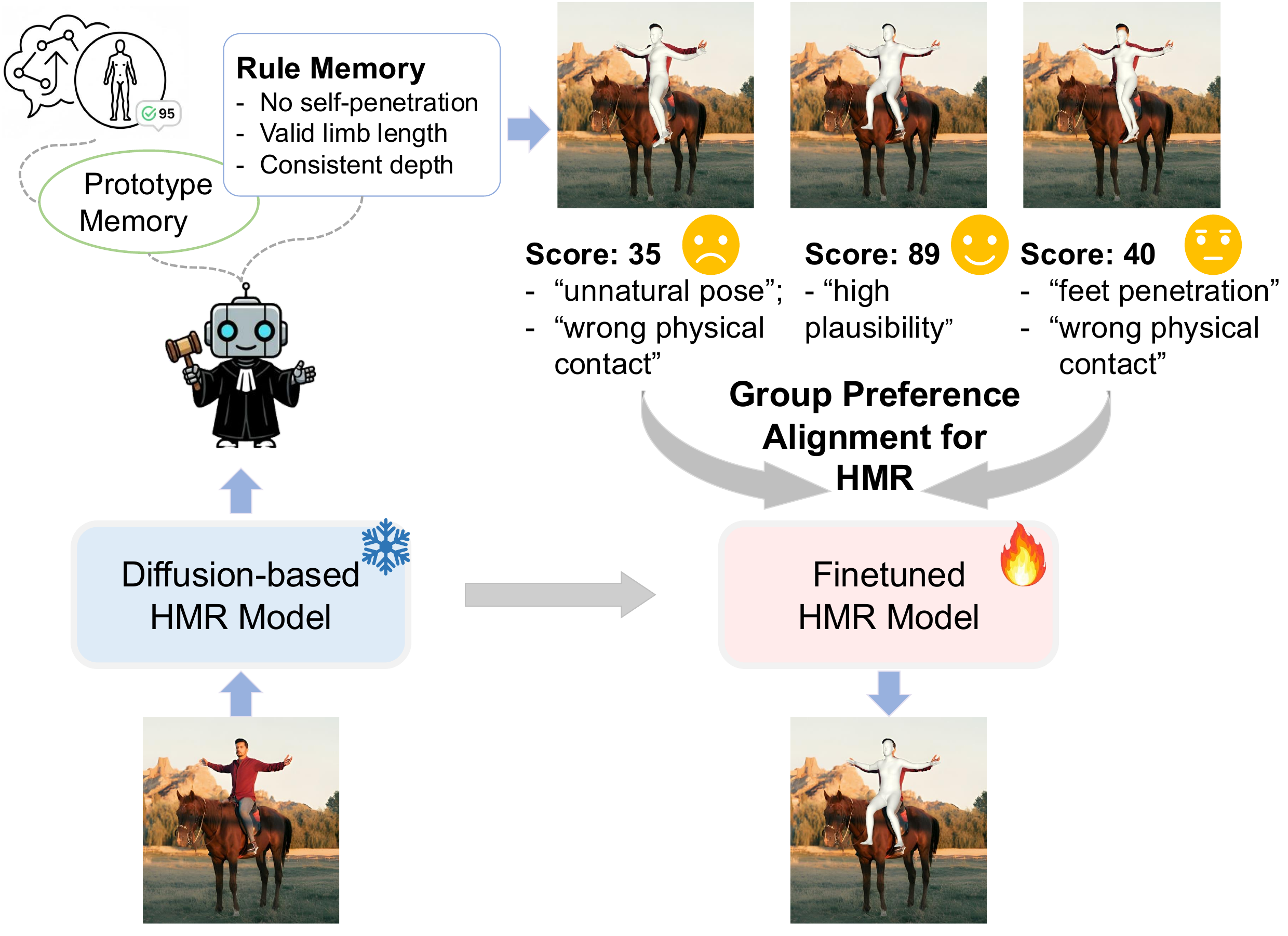}
    \vspace{-10pt}
    \caption{
    We introduce a VLM-guided HMR critique agent equipped with a dual-memory mechanism that delivers stable and semantically grounded assessments for groups of estimated 3D meshes. 
    Building on these group-wise signals, our group preference alignment framework steers diffusion-based HMR models towards more coherent and reliable mesh generation.
    \vspace{-10pt}
    }
\label{fig:teaser}
\end{figure}

It is observed that diffusion-based HMR methods~\cite{foo2023hmdiff, xu2024scorehypo} often produce 3D meshes inconsistent with the 2D input, due to the lack of explicit 3D joint re-projection constraints. ADHMR~\cite{shen2025adhmr} addresses this with Direct Preference Optimization (DPO), using a scorer to rank prediction quality. However, the image-driven scorer can be misled, favoring silhouette-aligned but physically implausible poses. 
In challenging scenarios, such as occlusion or incorrect contacts (e.g., floating feet, self-penetration), DPO often provides unreliable guidance. This is because DPO relies only on pairwise comparisons and ignores the quality relationships among multiple predictions.

In this work, we propose a comprehensive framework that utilizes a Visual Language Model (VLM)-based critique agent to improve the performance of the diffusion-based HMR model, as illustrated in Fig.~\ref{fig:teaser}.
We observe that large VLMs already encode strong priors about human pose semantics, contact relations, and spatial consistency, enabling them to assess the physical and geometric plausibility of 3D reconstructions from a single image. 
However, because of the criteria drift and subjective biases~\cite{shankar2024validates, li2025generation}, their raw judgments often lead to inconsistent or incomparable scores between prediction samples. 
To address this, we introduce a dual-memory augmented critique agent that can actively retrieve and reflect.
In the exploration phase, it performs self-reflection to mine new judging rules and typical prediction prototypes, thereby refining its reasoning.
During the evaluation phase, memory updates are disabled. The critique agent automatically retrieves relevant judging rules and prototype examples from its memory to ground each evaluation, yielding stable and semantically aligned assessments under occlusion or complex scene contexts.

Then, we aim to transfer the rich knowledge embedded in the critique agent to the human mesh estimator.
Inspired by the success of Group Relative Policy Optimization (GRPO)~\cite{shao2024deepseekmath, guo2025deepseekr1} in aligning large language models, we adapt its group-wise preference learning to diffusion-based HMR.
However, applying GRPO to diffusion is challenging because GRPO relies on stochastic rollouts, whereas diffusion models typically rely on deterministic ODE samplers for better efficiency and quality~\cite{song2020ddim, lu2022dpm}. 
Prior attempts~\cite{xue2025dancegrpo, liu2025flowgrpo} introduce stochasticity via SDE sampling but require training along the entire diffusion trajectory, increasing computation cost and reducing output fidelity.

We keep the efficiency of ODE-based diffusion and focus on the key benefit of GRPO, which is learning from group-level preference signals. 
Specifically, for each image, we generate several mesh hypotheses, score them with the critique agent, and convert these scores into advantages that capture their relative quality. 
Using these advantages, we formulate a simple and ODE-compatible group preference loss that transfers the critique agent’s judgment without requiring trajectory-level reinforcement.
We find that this group-wise, advantage-weighted alignment enables the model to discern and refine subtle pose details, producing more semantically consistent results. 
Crucially, because our method relies on relative preference signals from the critique model, it remains effective even when finetuned on in-the-wild datasets with unreliable 3D annotations.

To summarize, our main contributions are:
\begin{itemize}
    \item We propose a dual-memory augmented, self-reflective HMR critique agent to ensure consistent, semantically grounded scoring for human mesh predictions.
    \item We propose a group preference alignment framework for diffusion-based HMR models that operates without 3D ground truth. This design enables effective finetuning on noisy in-the-wild datasets and leads to more plausible and better 2D-aligned human mesh predictions.
    \item Extensive experiments show that this framework can effectively improve diffusion-based HMR models on challenging in-the-wild datasets.
\end{itemize}

\section{Related Work}
\noindent \textbf{Human Mesh Recovery from a Single Image.}
Current HMR methods can be broadly divided into deterministic and probabilistic paradigms.
Deterministic methods produce a single estimate for each input image. 
Early optimization-based approaches~\cite{bogo2016keepitsmpl, lassner2017unite, pavlakos2019smplxehf, joo2021eft} are time-consuming and often trapped in local minima due to poor initialization. 
More recent works~\cite{kanazawa2018hmr, li2021hybrik, goel2023hmr2.0, dwivedi2024tokenhmr, sarandi2024nlf, patel2025camerahmr, wang2025prompthmr} leverage deep neural networks to directly regress human model parameters, achieving strong performance yet still failing to address inherent depth and occlusion ambiguities in monocular settings.
To tackle these ambiguities, probabilistic methods~\cite{kolotouros2021prohmr, fang2023posterior, biggs20203dmultibodies, xu2024scorehypo, shen2025adhmr} aim to generate multiple predictions or model full posterior distributions. 
ScoreHypo~\cite{xu2024scorehypo} introduces a diffusion-based generator to produce diverse hypotheses and relies on an auxiliary selector network to choose the best estimate. 
ADHMR~\cite{shen2025adhmr} uses Diffusion-DPO to align diffusion sampling with a learned HMR-Scorer. However, its 2D-based scorer is easily misled by occlusions and cluttered backgrounds. 
MEGA~\cite{fiche2025mega} explores stochastic generation through iterative token unmasking but remains limited by the expressiveness of its vector-quantized mesh codebook.
In contrast, we extend GRPO to diffusion in an offline setting, using group-wise advantages to refine sampling and enhance fidelity.

\begin{figure*}
    \centering
    \includegraphics[width=1\linewidth]{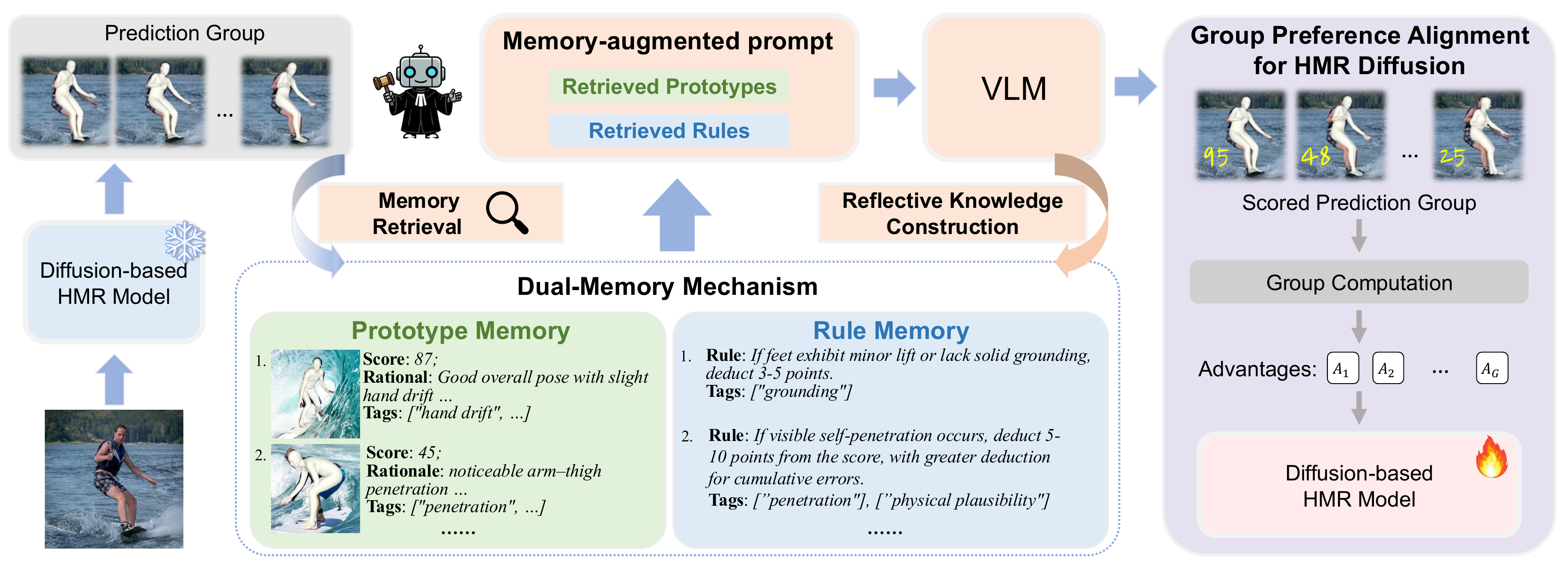}
    \vspace{-20pt}
    \caption{
    Overview of our framework.
    Our purpose is to refine a diffusion-based HMR model that generates a group of human mesh predictions per input image.
    We propose a VLM-enhanced HMR critique agent that assigns a score for each human mesh prediction. This critique agent is equipped with a dual-memory mechanism to give stable assessments.
    Then, we use this critique agent to build a group-wise HMR preference dataset without the need for manual labeling.
    Finally, we employ this preference dataset to finetune the base model to preferentially generate predictions that are physically plausible and better aligned with the image cues.
    \vspace{-10pt}
    }
\label{fig:overview}
\end{figure*}

\noindent \textbf{Preference Learning for Diffusion Models.}
LLM alignment commonly uses Reinforcement Learning from Human Feedback (RLHF)~\cite{christiano2017rlhf}, but it suffers from high cost and instability. While Direct Preference Optimization (DPO)~\cite{rafailov2024dpo} is simpler, it is often weaker. Recently, Group Relative Policy Optimization (GRPO)~\cite{shao2024deepseekmath, guo2025deepseekr1} has emerged as a more stable and powerful framework, markedly improving LLM reasoning.
Inspired by this success, recent studies have begun exploring aligning diffusion models. 
Early works~\cite{xu2024imagereward, fan2023dpok, black2023ddpo, clark2024draft} follow the RLHF paradigm, while later efforts adapt DPO to diffusion~\cite{wallace2024diffusiondpo, yang2024d3po, liang2025spo, li2024diffusionkto}.
Only a few attempts explore GRPO for diffusion, hindered by the mismatch between GRPO’s stochastic exploration and the deterministic nature of ODE-based diffusion sampling.
Recent works~\cite{xue2025dancegrpo, liu2025flowgrpo} address this via SDE sampling, but at the cost of higher computation and slower convergence.
Our method extends GRPO to diffusion in an offline setting, retaining both the stability of group optimization and the efficiency of standard diffusion models.

\noindent \textbf{LLM as a Scorer.}
LLMs are increasingly used as scalable alternatives to expert judgment. Existing approaches include fine-tuned judges~\cite{huang2025empirical, wang2025trustjudge, zhu2023judgelm, kim2023prometheus}, prompted judges~\cite{li2024crowdsourced, qin2024large, dubois2023alpacafarm}, and multi-agent judges~\cite{chan2023chateval, zhang2025sentient, verga2024replacing}.
Fine-tuned models offer strong accuracy but often fail to generalize, prompted methods depend heavily on prompt design, and multi-agent pipelines are computationally expensive.
We introduce the first VLM-enhanced scoring agent for HMR, equipped with dual memory and self-reflection to retrieve judging rules and visual prototypes, enabling consistent and semantically grounded evaluation without finetuning.
\section{Method}

\subsection{Preliminaries}

\noindent \textbf{Diffusion Models.}
Diffusion models~\cite{ho2020ddpmdiffusion, song2021score} define a forward process that progressively adds Gaussian noise to clean data $\mathbf{x}$ over $T$ timesteps. 
In our setting, $\mathbf{x}$ represents the joint swing and twist parameters following~\cite{xu2024scorehypo, li2021hybrik}, which are later converted into SMPL~\cite{loper2015smpl} pose parameters.
The forward process follows
$q(\mathbf{x}_t|\mathbf{x}_{t-1}) = \mathcal{N}\!\big(\sqrt{1-\beta_t}\,\mathbf{x}_{t-1},\, \beta_t \mathbf{I}\big)$,
where the noisy sample can be expressed as 
$\mathbf{x}_t = \sqrt{\bar{\alpha}_t}\mathbf{x}_0 + \sqrt{1-\bar{\alpha}_t}\boldsymbol{\epsilon}$, 
with $\boldsymbol{\epsilon}\!\sim\!\mathcal{N}(0,\mathbf{I})$.
The reverse process is modeled as
\begin{equation}
    p_\theta(\mathbf{x}_{t-1}|\mathbf{x}_t) = \mathcal{N}\!\big(\mathbf{x}_{t-1};\, \mu_\theta(\mathbf{x}_t,t),\, \sigma_t^2 \mathbf{I}\big),
    \label{eq:diffusion}
\end{equation}
where $\beta_t$, $\alpha_t$ and $\sigma_t$ are the noise schedules, $\mu_\theta$ is computed from the noise prediction model $\boldsymbol{\epsilon}_\theta$.
The model is trained by minimizing a denoising loss:
\begin{equation}
    L_{\text{DM}} = \mathbb{E}_{t,\mathbf{x}_0,\boldsymbol{\epsilon}}\left[ \lambda_t \, \| \boldsymbol{\epsilon} - \boldsymbol{\epsilon}_\theta(\mathbf{x}_t, t) \|_2^2 \right],
    \label{eq:diffusion_loss}
\end{equation}
where $\lambda_t$ is a timestep weighting function.
In the continuous-time view~\cite{song2021score}, the diffusion process is modeled as a stochastic differential equation (SDE), and its reverse-time SDE governs denoising. Removing the stochastic term yields a deterministic probability-flow ODE, which DDIM~\cite{song2020ddim} leverages for faster, consistent sampling with the same trained model.

\noindent \textbf{Group Relative Policy Optimization (GRPO).}
GRPO~\cite{shao2024deepseekmath} extends Reinforcement Learning from Human Feedback (RLHF) to a group-wise formulation, providing more stable policy updates while removing the need for a separate value model.
For each condition $\boldsymbol{c}$, a set of $G$ samples $\{\mathbf{x}_0^{i}\}_{i=1}^G$ is generated from the old policy $p_{\theta_{\text{old}}}(\cdot|\boldsymbol{c})$.
Formally, GRPO updates the policy parameters $\theta$ by maximizing the advantage-weighted likelihood ratio (we omit the clip term and KL term hereinafter for brevity): 
\begin{equation} 
\max_{\theta}\; \mathbb{E}_{\boldsymbol{c},\, \{\mathbf{x}_0^{i}\}}\! \left[ \sum_{i=1}^{G} \frac{p_{\theta}(\mathbf{x}_0^{i}|\boldsymbol{c})} {p_{\text{ref}}(\mathbf{x}_0^{i}|\boldsymbol{c})} \, A(\mathbf{x}_0^{i}) \right], 
\label{eq:grpo-core} 
\end{equation} 
where the advantage term is computed using a group of rewards $\{r_{i}\}_{i=1}^G$ corresponding to samples within the group:
\begin{equation}
    A_i = \frac{r_i - \mathrm{mean}(\{r_1, r_2, \ldots, r_G\})}{\mathrm{std}(\{r_1, r_2, \ldots, r_G\})}.
\label{eq:grpo-advantage}
\end{equation}

\subsection{Overview}
An overview of our proposed framework is presented in Fig.~\ref{fig:overview}.
Given an input image $I$, we aim to predict the human pose parameters ${\theta} \in \mathbb{R}^{24 \times 3}$ and shape parameters ${\beta} \in \mathbb{R}^{10}$ of the SMPL model~\cite{pavlakos2019smplxehf}. 
Following \cite{xu2024scorehypo}, we formulate this problem using a reverse diffusion process conditioned on the input image features $\boldsymbol{c}$ to tackle the inherent reconstruction ambiguity.
We first propose a VLM-guided critique agent to assess the quality of human mesh predictions given an input image in Sec.~\ref{sec:agent}. 
Next, we construct a synthetic group-wise HMR preference dataset (Sec.~\ref{sec:grodataset}). Candidate human mesh predictions are generated by the reference model $\boldsymbol{\epsilon}_{\text{ref}}$ and then grouped based on scores provided by the critique agent.
To distill knowledge from this synthetic preference dataset, we propose a group-wise preference optimization framework that finetunes an HMR diffusion model $\boldsymbol{\epsilon}_{\theta}$ to preferentially generate preferred human mesh predictions (Sec.~\ref{sec:diffgrpo}), and $\boldsymbol{\epsilon}_{\theta}$ is initialized from $\boldsymbol{\epsilon}_{\text{ref}}$.

\subsection{VLM-Guided HMR Critique Agent}
\label{sec:agent}

To automate the nuanced task of evaluating HMR predictions, we introduce a VLM-guided HMR critique agent $\mathcal{C}_{\text{VLM}}$. 
Unlike traditional regressors that predict a score from 2D joint data~\cite{shen2025adhmr}, our agent emulates expert human judgment by working directly from rendered images. 
Formally, given an RGB image $I$ and $n$ human mesh prediction overlays $\{\hat I_j\}_{j=1}^{n}$ rendered on $I$, the agent outputs per-overlay scores $s_j\in[0,100]$ and one-sentence critiques $c_j$.

The agent's architecture combines a VLM scoring engine with a dual-memory mechanism and a self-reflective knowledge construction loop. It operates in two phases: an exploration phase to learn and build its memory, and an evaluation phase where this frozen knowledge is applied for consistent, robust scoring.

\subsubsection{Dual-Memory Mechanism}
The foundation of the agent is a VLM scoring engine guided by a dual-memory augmentation design.
This design provides a lightweight yet interpretable prior that can be queried and expanded over time.
Specifically, we use two different types of memory:
\begin{itemize}
    \item \textbf{Rule Memory}: $\mathcal{M}_R = \{(t_i, T_i, N^u_i, N^s_i)\}_{i=1}^R$, where $t_i$ is the assessment rule text, $T_i$ is a set of semantic tags, $N^u_i$ is a use count (tracking how often the rule has been applied), and $N^s_i$ is a success count (tracking how often its application led to a ground-truth-aligned score).
    \item \textbf{Prototype Memory}: $\mathcal{M}_P$, stores previously judged hypothesis images as precedents. $\mathcal{M}_P = \{(v_i, r_i, T_i)\}_{i=1}^P$, where $v_i$ is a visual embedding of the SMPL prediction overlay image using CLIP~\cite{clip}, and $r_i$ is the textual rationale (including assigned score) for its assessment. $T_i$ is a set of semantic tags.
\end{itemize}

\noindent
\textbf{Dual-memory augmented scoring.}
The dual-memory augmented scoring process is the agent's core action, utilized in both the exploration and evaluation phases. It consists of three steps: 
(1) The agent queries prototype memory $\mathcal{M}_P$ to retrieve the top-$K$ most visually similar past examples by maximizing cosine similarity with the query image embedding $v_q$. 
(2) The agent queries $\mathcal{M}_R$ using a hybrid score $\Psi_j$ to select the most effective rules. This score combines semantic relevance ($\mathrm{R}$) with UCB exploration ($\mathrm{U}$):
\begin{equation}
    \Psi_i = \mathrm{R}(T_q, T_i) + \mathrm{U}_i.
\end{equation}
The Relevance Score, $\mathrm{R}(T_q, T_i) = |T_q \cap T_i|$, rewards rules matching the query tags $T_q$ from retrieved prototypes.
The Upper Confidence Bound (UCB)~\cite{auer2002ucb1} score is:
\begin{equation}
    \mathrm{U}_i = \rho_i + C \sqrt{\frac{\log N_{\text{total}}}{N^u_i + 1}},
\end{equation}
which balances the exploitation of high-success-rate rules and the exploration of rarely used rules.
Here, $\rho_i = N^s_i / N^u_i$ is the historical success rate, $C$ is an exploration constant, and $N_{\text{total}} = 1 + \sum_i N^u_i$ is the total number of rule applications.
(3) Finally, a context-rich prompt is dynamically constructed, including the retrieved rules and prototype rationales. This prompt is fed to the VLM, which generates the final score $s$ and critique $c$.

\subsubsection{Reflective Knowledge Construction}

We observe that while VLMs possess an inherent ability to judge 3D human mesh quality, direct prompting often yields unstable and inconsistent results. To this end, we allow the agent to build and refine domain-specific human pose estimation knowledge by itself in the exploration phase to stabilize its scoring capability.

The exploration phase is an iterative process where the agent's primary goal is to populate its dual-memory by learning from ground-truth (GT) data: 
(1) The agent first performs the memory-augmented scoring process on a batch of data, which increments the use count $N^u_i$ for every rule applied. 
(2) After scoring, it performs prototype memory write-back, saving salient examples back into $\mathcal{M}_P$. 
(3) The agent then performs an update for existing rules. 
It compares its score ranking against GT metrics via Spearman rank correlation.
If the correlation is higher than a threshold $\tau$, the applied rules are deemed effective, and their corresponding success count $N^s_i$ is incremented, reinforcing their efficacy. 
(4) The core learning step is new rules mining via reflection. The agent instructs the VLM to inspect the discrepancy between its own output and the GT metrics and then propose 1-2 new, testable rules. 
These auto-mined rules are added to $\mathcal{M}_R$ as new entries.

In the evaluation phase, the agent's learning loop and memory are frozen. It exclusively performs the memory-augmented scoring to generate consistent, final evaluations on unseen data.

\subsection{HMR Group Preference Dataset}
\label{sec:grodataset}
To effectively train our mesh prediction policy using GRPO, we must first construct a dataset that provides richer supervision than standard pairwise preferences. 
However, manually annotating $G$ complex 3D meshes is prohibitively slow, expensive, and subject to high inter-annotator variance. 
We overcome this bottleneck by leveraging our critique agent $\mathcal{C}_{\text{VLM}}$ as a high-fidelity, automated annotator.

The dataset construction process follows two main stages:
(1) \textbf{Group generation}: For each image $I$ in our training corpus, we first require a diverse set of $G$ human mesh predictions, $\{\mathbf{m}^i\}_{i=1}^G$.
This set spans a wide range of qualities, from common failure modes to high-quality alignments. 
We generate this group by sampling our frozen, pre-trained diffusion-based HMR model, $\boldsymbol{\epsilon}_{\text{ref}}$ for $G$ times, conditioned on $I$ but initiated with different initial noises. 
(2) \textbf{Group-wise scoring}: 
We employ our critique agent $\mathcal{C}_{\text{VLM}}$ to score the entire group of $G$ predictions simultaneously to obtain consistent relative quality judgments within one group. 
The model is prompted with the image $I$ and the complete set of mesh predictions $\{\mathbf{m}^i\}_{i=1}^G$ (visualized as 2D overlays) to get a set of scalar quality scores $\{s^i\}_{i=1}^G $:
\begin{equation}
    \{s^1, \ldots, s^G\} = \mathcal{C}_{\text{VLM}}(I, \mathbf{m}^1, \ldots, \mathbf{m}^G).
\end{equation}
By applying the scoring to all generated groups, we obtain our final, fully annotated group-wise preference dataset, $\mathcal{G}_{\text{HMR}}$. Each entry in this dataset is a tuple containing the input image, the group of predictions, and their corresponding VLM-assigned scores $\mathcal{G}_{\text{HMR}} = \left\{ \left( I, (\mathbf{m}^1, s^1), (\mathbf{m}^2, s^2), \ldots, (\mathbf{m}^G, s^G) \right) \right\}$.
This collection provides the dense preference signals $\{s^i\}$ required to train our GRPO refinement policy.

\subsection{Group Preference Alignment for HMR}
\label{sec:diffgrpo}

Once we obtain the group-wise preference dataset $\mathcal{G}_{\text{HMR}}$ that encodes rich quality cues for HMR, we seek to instill this rich preference information into our diffusion model. 
While pairwise methods like Diffusion-DPO~\cite{wallace2024diffusiondpo, shen2025adhmr} exist, probabilistic HMR models inherently generate multiple hypotheses to address the one-to-many mapping ambiguity. This multi-hypothesis approach necessitates a learning signal that is more stable and comprehensive than comparisons between isolated pairs.
To this end, we introduce group preference alignment for HMR diffusion models.
Our objective is to optimize a diffusion-based HMR estimator $\boldsymbol{\epsilon}_\theta$, initialized from a reference denoiser $\boldsymbol{\epsilon}_\text{ref}$, such that it learns from the static group-wise HMR preference data and produces human mesh predictions that are more physically plausible and better aligned with image evidence.

Concretely, given the human mesh group preference dataset $\mathcal{G}_{\text{HMR}}$ and the image condition $\boldsymbol{c}$, we follow GRPO (Eq.~(\ref{eq:grpo-advantage})) to compute the relative prediction quality advantage $A(\cdot)$ using their quality scores ${s^1, \ldots, s^G}$:
\begin{equation}
    A_i = \frac{s_i - \mathrm{mean}(\{s_i\}_\text{i=1}^G)}{\mathrm{std}(\{s_i\}_\text{i=1}^G)}.
\label{eq:G1}
\end{equation} 
Building on the GRPO objective in Eq.~(\ref{eq:grpo-core}), which maximizes advantage-weighted likelihood ratios within each group, we treat the diffusion sampler as a conditional policy $p_\theta(\mathbf{m}\mid\boldsymbol{c})$ parameterized by the noise predictor $\boldsymbol{\epsilon}_\theta$.
Inspired by \cite{rafailov2024dpo, wallace2024diffusiondpo}, we then instantiate the group-wise update rule as an advantage-weighted log-likelihood ratio between the trainable policy $p_\theta$ and the frozen reference policy $p_{\mathrm{ref}}$:
\begin{equation}
\mathcal{L}(\theta)
=
-\mathbb{E}_{\boldsymbol{c},\,\{\mathbf{m}^i\}}
\left[
\sum_{i=1}^{G}
A(\mathbf{m}^i)\,
\log
\frac{p_{\theta}(\mathbf{m}^i\mid\boldsymbol{c})}
     {p_{\mathrm{ref}}(\mathbf{m}^i\mid\boldsymbol{c})}
\right].
\label{eq:gpa-logratio}
\end{equation}

To specialize this objective to diffusion-based HMR, we follow the reparameterization used in Diffusion-DPO~\cite{wallace2024diffusiondpo} and express the log-ratio through the corresponding diffusion paths
$\mathbf{x}_{0:T}^i$:
\begin{equation}
\log
\frac{p_{\theta}(\mathbf{m}^i\mid\boldsymbol{c})}
     {p_{\mathrm{ref}}(\mathbf{m}^i\mid\boldsymbol{c})}
\;\approx\;
\mathbb{E}_{q(\mathbf{x}_{1:T}^i\mid\mathbf{m}^i)}
\log
\frac{p_{\theta}(\mathbf{x}_{0:T}^i)}
     {p_{\mathrm{ref}}(\mathbf{x}_{0:T}^i)},
\label{eq:path-ratio-simple}
\end{equation}
where we approximate the intractable posterior over paths with the forward
noising process $q(\mathbf{x}_{1:T}^i\mid\mathbf{m}^i)$.
Under the standard Gaussian parameterization of the reverse diffusion process (Eq.~\eqref{eq:diffusion}), $p_\theta$ is induced by the noise predictor $\boldsymbol{\epsilon}_\theta$, and the log-likelihood ratio admits the following diffusion surrogate (we omit $\boldsymbol{c}$ for brevity hereinafter):
\begin{equation}
    \log
    \frac{p_{\theta}(\mathbf{m}^i\mid\boldsymbol{c})}
         {p_{\mathrm{ref}}(\mathbf{m}^i\mid\boldsymbol{c})}
    \;\approx\;
    T\lambda_t\,\mathbb{E}_{t,\boldsymbol{\epsilon}}
    \Big[
    L_{\mathrm{DM}}^{\mathrm{ref}}(\mathbf{x}_t^i,\boldsymbol{\epsilon})
    -
    L_{\mathrm{DM}}^{\theta}(\mathbf{x}_t^i,\boldsymbol{\epsilon})
    \Big],
\end{equation}
where $t\sim\mathcal{U}(\{1,\ldots,T\})$, $\boldsymbol{\epsilon}\sim\mathcal{N}(0,\mathbf{I})$, $\lambda_t$ is a timestep weighting function,  $\mathbf{x}_t^i$ is obtained by diffusing $\mathbf{m}^i$, and
\begin{equation}
L_{\mathrm{DM}}^{\theta}(\mathbf{x}_t^i,\boldsymbol{\epsilon})
=
\big\|
\boldsymbol{\epsilon}_{\theta}(\mathbf{x}_t^i,t)
-
\boldsymbol{\epsilon}
\big\|_2^2
\end{equation}
is the standard diffusion noise prediction loss (with an analogous definition for
$L_{\mathrm{DM}}^{\mathrm{ref}}$).
Plugging this surrogate into~\eqref{eq:gpa-logratio} yields our final
training objective:
\begin{equation}
    \begin{gathered}
        \mathcal{L}(\theta) = \mathbb{E}_{\mathbf{m} \sim \mathcal{G}_\text{HMR}, t \sim \mathcal{U}(1, T),\boldsymbol{\epsilon}\!\sim\!\mathcal{N}(0,\mathbf{I})} \\
         \beta T \lambda_t 
       \sum_{i=1}^G\!  \Big[ A(\mathbf{m}^i)\,
        (L_{\text{DM}}^{\mathrm{\theta}}(\mathbf{x}_t^i,\boldsymbol{\epsilon})\!
        - \!L_{\text{DM}}^{\mathrm{ref}}(\mathbf{x}_t^i,\boldsymbol{\epsilon}))  \Big], 
    \end{gathered}
\label{eq:loss_final}
\end{equation}
where the hyperparameter $\beta$ controls regularization.

Intuitively, meshes with higher critique scores (positive advantages) are encouraged to achieve a smaller denoising loss than the reference model, while lower-scored meshes are pushed in the opposite direction, aligning the HMR diffusion model with the group-wise preference signal.

\section{Experiments}
\label{sec:exp}

\begin{table*}
    \centering
    \resizebox{0.87 \linewidth}{!}
    {
    \begin{tabular}{c|l|c|ccc|ccc}
    \toprule
    & \multirow{2}{*}{\rule{0pt}{3.0ex}Methods } & \multirow{2}{*}{\rule{0pt}{3.0ex}$M$} & \multicolumn{3}{c|}{3DPW~\cite{3dpw}} & \multicolumn{3}{c}{Human3.6M~\cite{human36m}}   \\
     \cline{4-9}
    & & & {\rule{0pt}{2.7ex} PVE $\downarrow$ } & { MPJPE $\downarrow$ } & { PA-MPJPE $\downarrow$ } & { PVE $\downarrow$ } & { MPJPE $\downarrow$ } & { PA-MPJPE $\downarrow$ } \\
    \midrule
    \multirow{9}{*}{\parbox{0.5cm}{\centering\rotatebox[origin=c]{90}{\textbf{Deterministic}}}}
    & HMR~\cite{kanazawa2018hmr}        & - & 152.7 & 130.0 & 81.3 & 96.1 & 88.0 & 56.8 \\
    & HybrIK~\cite{li2021hybrik}       & - & 86.5  & 74.1  & 45.0 & 65.7 & 54.4 & 34.5 \\
    & PyMaf~\cite{zhang2021pymaf}         & -  & 110.1 & 92.8 & 58.9  & -    & 57.7 & 40.5 \\
    & HMR 2.0~\cite{goel2023hmr2.0} & - & -    & 70.0 & 44.5 & -  & 44.8 & 33.6 \\
    & POTTER~\cite{zheng2023potter}         & -  & 87.4  & 75.0 & 44.8 & -    & 56.5 & 35.1  \\
    & Zolly~\cite{wang2023zolly}            & -  & 76.3  & 65.0 & 39.8 & - & 49.4 & 32.3 \\
    & ScoreHMR~\cite{stathopoulos2024scorehmr} & - & - & 76.8 & 51.1 & -  & - & - \\
    & CameraHMR~\cite{patel2025camerahmr} & - & 65.9 & 56.0 & 35.1 & -  & - & -  \\
    & PromptHMR~\cite{wang2025prompthmr} & - & 69.4 & 58.7 & 36.6 & -  & - & -  \\
    \midrule
    \multirow{16}{*}{\parbox{0.5cm}{\centering\rotatebox[origin=c]{90}{\textbf{Probabilistic}}}}
    & 3D Multibodies~\cite{biggs20203dmultibodies} & 25  & -    & 75.8 & 55.6 & -    & 58.2 & 42.2 \\
    & ProHMR~\cite{kolotouros2021prohmr} & 25 & -    & -    & 52.4 & -    & -    & 36.8 \\
    & Sengupta \textit{et al.}~\cite{sengupta2021hierarchical} & 25    & -    & 75.1 & 47.0    & -    & -    & -   \\
    & HMDiff~\cite{foo2023hmdiff} & 25   & 82.4 & 72.7 & 44.5  & -    & 49.3 & 32.4 \\
    & HuManiFlow ~\cite{sengupta2023humaniflow} & 100 & -    & 65.1 & 39.9  & -    & -    & -  \\
    & \multirow{2}{*}{ScoreHypo~\cite{xu2024scorehypo}} & \rule{0pt}{2.4ex} 10 & 79.8 & 68.5 & 41.0 & 52.5 & 42.4 & 29.0  \\
    & & 100 & 73.4 & 63.0 & 37.6 & 47.5 & 38.4 & 26.0  \\
    & MEGA~\cite{fiche2025mega} & 25 & 75.1 & 63.6 & 40.4 & - & - & - \\
    & \multirow{2}{*}{ADHMR~\cite{shen2025adhmr} } & \rule{0pt}{2.4ex} 10 & 73.8 & 64.2 & 38.3 & 52.1 & 41.8 & 28.4 \\
     & & 100 & {65.4} & {57.2} & {33.5} & 45.9 & 36.9 & 24.8 \\
     \cline{2-9} 
    & \multirow{3}{*}{ \textbf{Ours} } & \rule{0pt}{2.4ex} 10 & 69.0 & 59.3 & 36.3 & 49.5 & 39.6 & 27.2 \\
     & & 100 & {60.9} & {52.5} & {31.5} & 43.8 & 35.0 & 23.9 \\
     & & 200 & 59.0 & {51.0} & \textbf{30.2} & {42.4} & 34.0 & {23.2} \\
    \cline{2-9} 
    & \multirow{3}{*}{ \textbf{Ours$^\dagger$} } & \rule{0pt}{2.4ex} 10 & 67.2 & 56.6 & 36.4 & 49.2 & 39.0 & 26.9 \\
     & & 100 & 59.5 & 49.9 & 31.9 & 43.2 & 34.3 & 23.5 \\
     & & 200 & \textbf{57.7} & \textbf{48.5} & {30.5} & \textbf{42.0} & \textbf{33.2} & \textbf{22.8} \\
    \bottomrule
    \end{tabular}
    }
    \vspace{-0pt}
    \caption{Comparison with state-of-the-arts on the 3DPW~\cite{3dpw} and Human3.6M~\cite{human36m} dataset. 
     $M$ is the number of predictions of probabilistic methods.
     Ours$^\dagger$ is trained on an additional in-the-wild dataset InstaVariety~\cite{kanazawa2019instavariety}, using only preference signals without 3D labels.
    }
    \label{tab:hmr}
\end{table*}

\begin{table}
    \centering
    \resizebox{0.94\linewidth}{!}
    {
    \begin{tabular}{l|ccc}
    \toprule
    \multirow{2}{*}{ Methods } & \multicolumn{3}{c}{ 3DPW~\cite{3dpw} }   \\
     \cline{2-4}
     & {\rule{0pt}{2.7ex}PVE $\downarrow$ } & { MPJPE $\downarrow$ } & { PA-MPJPE $\downarrow$ } \\
    \midrule
    Base Diffusion Model & 73.4 & 63.0 & 37.6   \\
    \hspace{1em}\textit{+ supervised finetuning} & 70.2 & 61.3 & 36.5  \\
    \midrule
    Ours & {\textbf{59.5}} & {\textbf{49.9}} & {\textbf{31.9}}  \\
    \hspace{1em}\textit{DPO w/ critique agent} & 63.9 & 53.1 & 33.4  \\
    \hspace{1em}\textit{w/o critique agent} & 65.4 & 54.9 & 34.7  \\
    \bottomrule
    \end{tabular}
    }
    \vspace{-5pt}
    \caption{Ablation study on the 3DPW~\cite{3dpw} test set. All models are finetuned on the InstaVariety~\cite{kanazawa2019instavariety} dataset. $M=100$ for all.
    }
    \vspace{-10pt}
    \label{tab:ablation1}
\end{table}

\begin{figure*}
    \centering
    \includegraphics[width=1.0\linewidth]{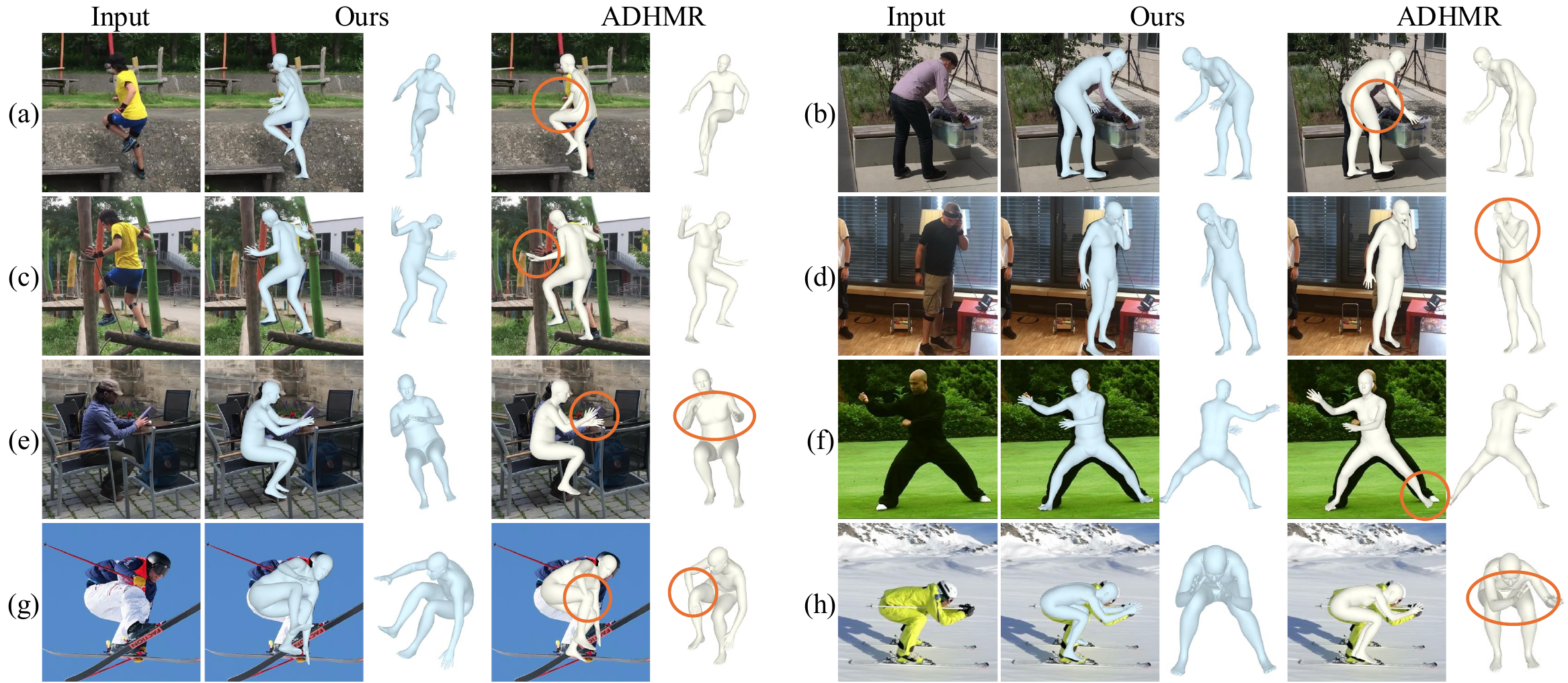}
    \vspace{-5pt}
    \caption{Qualitative comparison between our method and the state-of-the-art probabilistic model ADHMR~\cite{shen2025adhmr}. Examples (a) $\sim$ (e) are from the 3DPW~\cite{3dpw} dataset, while (f) $\sim$ (h) are challenging internet images. 
    Both overlay and side-view results are shown.
    }
    \vspace{-5pt}
\label{fig:qualitative}
\end{figure*}

\begin{figure*}
    \centering
    \includegraphics[width=1.0\linewidth]{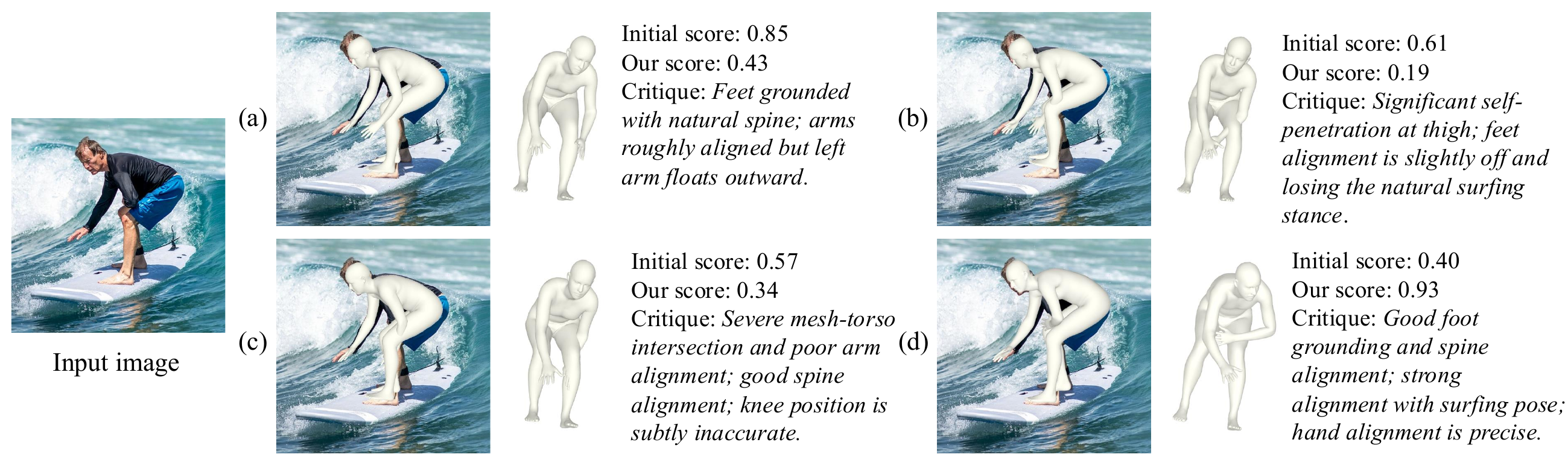}
    \vspace{-20pt}
    \caption{Visualization of our critique agent on an internet image. We compare the erroneous initial score from HMR-Scorer~\cite{shen2025adhmr} with the corrected score from our critique agent, along with the corresponding critique.}
    \vspace{-10pt}
\label{fig:agent}
\end{figure*}

\subsection{Experimental Setup}
\noindent
\textbf{Datasets.}
In the exploration phase, our critique agent samples images from five datasets: HI4D~\cite{yin2023hi4d}, BEDLAM~\cite{black2023bedlam}, DNA-Rendering~\cite{cheng2023dnarender}, GTA-Human II~\cite{cai2024gtahuman}, and SPEC~\cite{kocabas2021spec}.
These studio-based or synthetic datasets provide accurate 3D ground-truth annotations across diverse scenes, offering reliable quality signals for building the agent’s knowledge.
For the evaluation of the agent, we follow ADHMR~\cite{shen2025adhmr} to construct our test sets. This involves using data from the DNA-Rendering~\cite{cheng2023dnarender} and GTA-Human II~\cite{cai2024gtahuman} test splits.
For our reference diffusion model, we adopt the architecture from \cite{xu2024scorehypo, shen2025adhmr} and train it on standard HMR datasets: Human3.6M~\cite{human36m}, 3DPW~\cite{3dpw}, MPI-INF-3DHP~\cite{mehta20173dhp}, MPII~\cite{andriluka2014mpii}, COCO~\cite{lin2014mscoco}, and UP-3D~\cite{lassner2017unite}.

\noindent
\textbf{Metrics.}
We evaluate human mesh recovery using four standard metrics: PVE for mesh surface accuracy, MPJPE for 3D joint accuracy, and their Procrustes-aligned variants (PA-PVE and PA-MPJPE) that remove global misalignment.
To evaluate our critique agent, we follow standard score-prediction protocols~\cite{zhai2020iqasurvey, shen2025adhmr} and compute three correlation metrics between predicted scores and ground-truth quality: PLCC for linear correlation, and SRCC and KRCC for rank-based consistency. Higher values (closer to 1.0) indicate better correlation.

\noindent
\textbf{Implementation details.}
We use Qwen3-VL-32B as our VLM in the critique agent.
We use a learning rate of $1e^\text{-4}$ for the group preference finetuning with a batch size of 80 images. We use a group size of $G=20$ for training.

\subsection{Human Mesh Recovery Results}
\textbf{Quantitative comparison.}
Tab.~\ref{tab:hmr} shows the comparison of our method with state-of-the-art methods on two widely used benchmark datasets: Human3.6M~\cite{human36m} and in-the-wild 3DPW~\cite{3dpw}.
Following standard conventions for probabilistic methods~\cite{xu2024scorehypo, shen2025adhmr, biggs20203dmultibodies}, we generate $M$ estimates and report the minimum PVE, MPJPE, and PA-MPJPE.
When fine-tuned directly on these benchmarks, our method significantly improves the reference model's performance, demonstrating its ability to effectively leverage group-wise reward signals. Notably, on the 3DPW in-the-wild benchmark, our model achieves an 8.2\% MPJPE improvement ($M=100$) over ADHMR~\cite{shen2025adhmr}, even without using an additional in-the-wild dataset.
To further evaluate in-the-wild effectiveness, we introduce a variant, Ours$^\dagger$, which is fine-tuned on the InstaVariety~\cite{kanazawa2019instavariety} dataset. Crucially, instead of using the dataset's noisy pseudo 3D labels for a loss, we use its images to construct a new group-wise preference dataset via our critique agent and the frozen reference model. This approach allows our model to achieve superior results, surpassing the previous SOTA ADHMR~\cite{shen2025adhmr}. This performance boost highlights the generalizability of our critique agent and the effectiveness of our group preference alignment framework in enhancing in-the-wild generalization.

\begin{table*}
\centering
    \resizebox{1.0\linewidth}{!}
    {
    \begin{tabular}{l|cc|cc|cc|cc|cc|cc}
    \toprule
    \multirow{3}{*}{\rule{0pt}{8.0ex}Methods }  & \multicolumn{6}{c|}{ GTA-Human II~\cite{cai2024gtahuman} } & \multicolumn{6}{c}{ DNA-Rendering~\cite{cheng2023dnarender} }   \\
     \cmidrule{2-13}
    & \multicolumn{2}{c|}{ PVE } & \multicolumn{2}{c|}{ MPJPE } & \multicolumn{2}{c|}{ PA-MPJPE } & \multicolumn{2}{c|}{ PVE } & \multicolumn{2}{c|}{ MPJPE} & \multicolumn{2}{c}{ PA-MPJPE } \\
    \cmidrule{2-13}  & SRCC $\uparrow$ & KRCC $\uparrow$  & SRCC $\uparrow$ & KRCC $\uparrow$ & SRCC $\uparrow$ & KRCC $\uparrow$ & SRCC $\uparrow$ & KRCC $\uparrow$ & SRCC $\uparrow$ & KRCC $\uparrow$ & SRCC $\uparrow$ & KRCC $\uparrow$  \\
    \midrule 
    ScoreNet~\cite{xu2024scorehypo} & 0.464 & 0.411 & 0.468 & 0.404 & 0.430 & 0.327 & 0.532 & 0.512 & 0.496 & 0.469 & 0.458 & 0.367  \\
    HMR-Scorer~\cite{shen2025adhmr} &{0.578} & {0.506} & {0.588} & {0.513} & {0.489} & {0.428} & {0.610} & {0.540} & {0.562} & {0.508} & {0.475} & {0.393} \\
    \textbf{Ours} & \textbf{0.605} & \textbf{0.539} & \textbf{0.615} & \textbf{0.556} & \textbf{0.528} & \textbf{0.461} & \textbf{0.633} & \textbf{0.556} & \textbf{0.588} & \textbf{0.532} & \textbf{0.510} & \textbf{0.433}  \\
    \hspace{1em}\textit{w/o rule memory} & 0.597 & 0.491 & 0.577 & 0.480 & 0.454 & 0.427 & 0.619 & 0.487 & 0.564 & 0.520 & 0.484 & 0.377 \\
    \hspace{1em}\textit{w/o prototype memory} & 0.590 & 0.455 & 0.602 & 0.479 & 0.464 & 0.398 & 0.622 & 0.542 & 0.564 & 0.513 & 0.482 & 0.378 \\
    \hspace{1em}\textit{w/o self-reflection} & 0.534 & 0.426 & 0.542 & 0.444 & 0.440 & 0.285 & 0.568 & 0.463 & 0.567 & 0.460 & 0.483 & 0.298 \\
    \bottomrule
    \end{tabular}
    }
    \vspace{-5pt}
    \caption{Group-wise score prediction results. We report the SRCC and KRCC between the predicted scores and the ground-truth HMR metrics. Our method consistently outperforms all baselines and ablation variants.
    }
    \vspace{-10pt}
    \label{tab:group-scorer}
\end{table*}

\begin{table}
\centering
    \resizebox{1.0\linewidth}{!}
    {
    \begin{tabular}{l|ccc|ccc}
    \toprule
    \multirow{2}{*}{\rule{0pt}{4ex}Methods } & \multicolumn{3}{c|}{ GTA-Human II~\cite{cai2024gtahuman} - PLCC } & \multicolumn{3}{c}{ DNA-Rendering~\cite{cheng2023dnarender} - PLCC }   \\
     \cmidrule{2-7}
    & PVE & MPJPE & PA-MPJPE & PVE & MPJPE & PA-MPJPE \\
    \midrule 
    ScoreNet~\cite{xu2024scorehypo} & 0.524 & 0.516 & 0.472 & 0.550 & 0.551 & 0.499 \\
    HMR-Scorer~\cite{shen2025adhmr} &{0.634} & {0.627} & {0.565} & {0.664} & {0.659} & {0.620} \\
    \textbf{Ours} & \textbf{0.695} & \textbf{0.697} & \textbf{0.653} & \textbf{0.737} & \textbf{0.717} & \textbf{0.700} \\
    \hspace{1em}\textit{w/o rule memory} & 0.683 & 0.672 & 0.630 & 0.701 & 0.673 & 0.650 \\
    \hspace{1em}\textit{w/o prototype memory} & 0.645 & 0.629 & 0.597 & 0.677 & 0.659 & 0.629 \\
    \hspace{1em}\textit{w/o self-reflection} & 0.610 & 0.613 & 0.540 & 0.642 & 0.656 & 0.581 \\
    \bottomrule
    \end{tabular}
    }
    \vspace{-10pt}
    \caption{Point-wise score prediction results. We report the PLCC between the predicted scores and the ground-truth metrics.
    }
    \vspace{-10pt}
    \label{tab:point-scorer}
\end{table}

\noindent
\textbf{Qualitative results.}
Fig.~\ref{fig:qualitative} compares our method with the state-of-the-art probabilistic method ADHMR on challenging samples from 3DPW and in-the-wild internet images. The results prove our model's superior performance under severe occlusion and complex human-object interactions.
In cases (a) and (b), our model recovers more plausible poses for the partially occluded arms and achieves better alignment with the input image. In (d), where the subject is making a phone call, ADHMR misestimates the depth relationship between the left arm and the head, whereas our model correctly places the arm near the face, reflecting stronger semantic and spatial reasoning.
In (g), ADHMR generates self-penetration artifacts, while our result remains physically valid. In (h), despite heavy occlusion on the left side, our model still predicts a coherent 3D pose, whereas ADHMR falls into depth ambiguity.
These results collectively demonstrate that our model is substantially more robust than ADHMR on challenging in-the-wild scenarios.

\noindent
\textbf{Ablation study.}
Tab.~\ref{tab:ablation1} presents our ablation study. All variants are fine-tuned on the in-the-wild InstaVariety~\cite{kanazawa2019instavariety} dataset and evaluated on the 3DPW~\cite{3dpw} test set. 
The base diffusion model serves as our reference model, and we note that conventional supervised fine-tuning on noisy pseudo-labels provides only limited improvement.
To assess the value of our group preference alignment framework, we compare it with a DPO-style pairwise variant following ADHMR~\cite{shen2025adhmr}, using the same critique agent for fairness. 
Compared with the DPO-based variant, our group preference alignment framework reduces the MPJPE error by 6.0\%, reflecting its stronger ability to leverage group-wise preference signals and mitigate the mapping ambiguity of probabilistic models.
When removing the critique agent and using only the group alignment framework (\textit{Ours w/o critique agent}), we rely on the HMR-Scorer~\cite{shen2025adhmr} to build the preference dataset. This variant improves over the baseline but still lags behind our full model, highlighting the value of high-fidelity signals provided by the critique agent. Overall, the ablation results indicate that our performance gains arise from the combination of the critique agent’s richer supervision and the strong alignment capability of our method.

\subsection{HMR Critique Agent Results}

\noindent
\textbf{Evaluation protocol.} 
To evaluate the critique agent, we use two protocols: point-wise and group-wise.
Point-wise evaluation assesses the agent's ability to predict the absolute quality of each single prediction, while group-wise evaluation tests its ability to rank the relative quality of $K$ predictions for the same image.
Our point-wise benchmark is identical to that of ADHMR, using single perturbed GT poses to simulate predictions. 
For group-wise evaluation, we generate $K$ distinct predictions per image and compute correlation within each group, reporting the mean across groups.
We calibrate the critique agent's outputs to the HMR-Scorer’s numeric range using a quadratic programming post-processing step that learns a linear scale-and-shift transformation.

\noindent
\textbf{Group-wise results.}
Tab.~\ref{tab:group-scorer} reports the group-wise score prediction results.
We compare against two baselines: ScoreNet~\cite{xu2024scorehypo} and HMR-Scorer~\cite{shen2025adhmr}, both of which rely merely on joint-guided image features.
Results show that our critique agent outperforms all baseline methods in both SRCC and KRCC metrics, indicating its stronger capability to capture subtle 3D pose variations.
We further conduct ablations by removing each core component.
{\textit{Ours w/o rule memory}} removes the agent's access to the learned rules, leading to weaker and less stable ratings. 
{\textit{Ours w/o prototype memory}} removes instance-level prediction prototypes, which also degrades performance.
{\textit{Ours w/o self-reflection}} eliminates reflective knowledge construction in the exploration phase. This variant causes the most significant performance drop across all metrics, strongly underscoring that the self-reflection process is a critical component for enhancing the ranking stability of the agent.

\noindent
\textbf{Point-wise results.}
Tab.~\ref{tab:point-scorer} presents the point-wise score prediction results. 
The ablations further validate our design: removing either the rule memory or the prototype memory causes a noticeable performance drop, showing that both general rules and instance-level prototypes are important for accurate quality prediction. The self-reflection mechanism is also essential, as its removal leads to the largest decline, confirming its role in producing reliable scores.

\noindent
\textbf{Visualization.}
Fig.~\ref{fig:agent} illustrates the clear advantages of our critique agent over the HMR-Scorer baseline. The baseline's ``Initial score" is unreliable and appears to lack 3D awareness: it assigns a high score to a flawed pose (a) and an incorrectly low score to a high-quality 3D mesh (d).
In contrast, our critique agent demonstrates a robust comprehension of the 3D action and geometry. It correctly identifies that poses (a) and (c) fail to capture the proper surfing stance, especially the incorrectly extended left arm. 
Our agent also identifies major 3D geometric failures such as severe self-penetration (b, c).
This shows our agent's feedback is grounded in a precise understanding of 3D human motion, allowing it to identify subtle errors, thus rectifying the base model's flawed assessments via our group preference optimization framework.

\section{Conclusion}
In this work, we introduce a VLM-driven critique agent and a group preference alignment framework that together align diffusion-based HMR.
By providing semantically grounded preferences and learning from them without 3D supervision, our method produces more accurate human meshes, especially in challenging in-the-wild settings.
{
    \small
    \bibliographystyle{ieeenat_fullname}
    \bibliography{main}

@String(ECCV= {Eur. Conf. Comput. Vis.})

@String(TOG= {ACM Trans. Graph.})

@String(ECCV  = {ECCV})

@String(TOG   = {ACM TOG})

@inproceedings{dwivedi2024tokenhmr,
  title={Tokenhmr: Advancing human mesh recovery with a tokenized pose representation},
  author={Dwivedi, Sai Kumar and Sun, Yu and Patel, Priyanka and Feng, Yao and Black, Michael J},
  booktitle={Proceedings of the IEEE/CVF Conference on Computer Vision and Pattern Recognition},
  pages={1323--1333},
  year={2024}
}

@inproceedings{stathopoulos2024scorehmr,
  title={Score-guided diffusion for 3d human recovery},
  author={Stathopoulos, Anastasis and Han, Ligong and Metaxas, Dimitris},
  booktitle={Proceedings of the IEEE/CVF Conference on Computer Vision and Pattern Recognition},
  pages={906--915},
  year={2024}
}

@inproceedings{pavlakos2019smplxehf,
  title={Expressive body capture: 3d hands, face, and body from a single image},
  author={Pavlakos, Georgios and Choutas, Vasileios and Ghorbani, Nima and Bolkart, Timo and Osman, Ahmed AA and Tzionas, Dimitrios and Black, Michael J},
  booktitle={Proceedings of the IEEE/CVF conference on computer vision and pattern recognition},
  pages={10975--10985},
  year={2019}
}

@inproceedings{goel2023hmr2.0,
  title={Humans in 4d: Reconstructing and tracking humans with transformers},
  author={Goel, Shubham and Pavlakos, Georgios and Rajasegaran, Jathushan and Kanazawa, Angjoo and Malik, Jitendra},
  booktitle={Proceedings of the IEEE/CVF International Conference on Computer Vision},
  pages={14783--14794},
  year={2023}
}

@article{loper2015smpl,
  title={SMPL: a skinned multi-person linear model},
  author={Loper, Matthew and Mahmood, Naureen and Romero, Javier and Pons-Moll, Gerard and Black, Michael J},
  journal={ACM Transactions on Graphics (TOG)},
  volume={34},
  number={6},
  pages={1--16},
  year={2015},
  publisher={ACM New York, NY, USA}
}

@inproceedings{3dpw,
  title={Recovering accurate 3d human pose in the wild using imus and a moving camera},
  author={Von Marcard, Timo and Henschel, Roberto and Black, Michael J and Rosenhahn, Bodo and Pons-Moll, Gerard},
  booktitle={Proceedings of the European conference on computer vision (ECCV)},
  pages={601--617},
  year={2018}
}

@inproceedings{black2023bedlam,
  title={Bedlam: A synthetic dataset of bodies exhibiting detailed lifelike animated motion},
  author={Black, Michael J and Patel, Priyanka and Tesch, Joachim and Yang, Jinlong},
  booktitle={Proceedings of the IEEE/CVF Conference on Computer Vision and Pattern Recognition},
  pages={8726--8737},
  year={2023}
}

@article{human36m,
  title={Human3. 6m: Large scale datasets and predictive methods for 3d human sensing in natural environments},
  author={Ionescu, Catalin and Papava, Dragos and Olaru, Vlad and Sminchisescu, Cristian},
  journal={IEEE transactions on pattern analysis and machine intelligence},
  volume={36},
  number={7},
  pages={1325--1339},
  year={2013},
  publisher={IEEE}
}

@inproceedings{clip,
  title={Learning transferable visual models from natural language supervision},
  author={Radford, Alec and Kim, Jong Wook and Hallacy, Chris and Ramesh, Aditya and Goh, Gabriel and Agarwal, Sandhini and Sastry, Girish and Askell, Amanda and Mishkin, Pamela and Clark, Jack and others},
  booktitle={International conference on machine learning},
  pages={8748--8763},
  year={2021},
  organization={PMLR}
}

@inproceedings{li2021hybrik,
  title={Hybrik: A hybrid analytical-neural inverse kinematics solution for 3d human pose and shape estimation},
  author={Li, Jiefeng and Xu, Chao and Chen, Zhicun and Bian, Siyuan and Yang, Lixin and Lu, Cewu},
  booktitle={Proceedings of the IEEE/CVF conference on computer vision and pattern recognition},
  pages={3383--3393},
  year={2021}
}

@inproceedings{zhang2021pymaf,
  title={Pymaf: 3d human pose and shape regression with pyramidal mesh alignment feedback loop},
  author={Zhang, Hongwen and Tian, Yating and Zhou, Xinchi and Ouyang, Wanli and Liu, Yebin and Wang, Limin and Sun, Zhenan},
  booktitle={Proceedings of the IEEE/CVF International Conference on Computer Vision},
  pages={11446--11456},
  year={2021}
}

@inproceedings{bogo2016keepitsmpl,
  title={Keep it SMPL: Automatic estimation of 3D human pose and shape from a single image},
  author={Bogo, Federica and Kanazawa, Angjoo and Lassner, Christoph and Gehler, Peter and Romero, Javier and Black, Michael J},
  booktitle={Computer Vision--ECCV 2016: 14th European Conference, Amsterdam, The Netherlands, October 11-14, 2016, Proceedings, Part V 14},
  pages={561--578},
  year={2016},
  organization={Springer}
}

@inproceedings{lassner2017unite,
  title={Unite the people: Closing the loop between 3d and 2d human representations},
  author={Lassner, Christoph and Romero, Javier and Kiefel, Martin and Bogo, Federica and Black, Michael J and Gehler, Peter V},
  booktitle={Proceedings of the IEEE conference on computer vision and pattern recognition},
  pages={6050--6059},
  year={2017}
}

@inproceedings{kanazawa2018hmr,
  title={End-to-end recovery of human shape and pose},
  author={Kanazawa, Angjoo and Black, Michael J and Jacobs, David W and Malik, Jitendra},
  booktitle={Proceedings of the IEEE conference on computer vision and pattern recognition},
  pages={7122--7131},
  year={2018}
}

@inproceedings{joo2021eft,
  title={Exemplar fine-tuning for 3d human model fitting towards in-the-wild 3d human pose estimation},
  author={Joo, Hanbyul and Neverova, Natalia and Vedaldi, Andrea},
  booktitle={2021 International Conference on 3D Vision (3DV)},
  pages={42--52},
  year={2021},
  organization={IEEE}
}

@inproceedings{kolotouros2021prohmr,
  title={Probabilistic modeling for human mesh recovery},
  author={Kolotouros, Nikos and Pavlakos, Georgios and Jayaraman, Dinesh and Daniilidis, Kostas},
  booktitle={Proceedings of the IEEE/CVF international conference on computer vision},
  pages={11605--11614},
  year={2021}
}

@inproceedings{fang2023posterior,
  title={Learning analytical posterior probability for human mesh recovery},
  author={Fang, Qi and Chen, Kang and Fan, Yinghui and Shuai, Qing and Li, Jiefeng and Zhang, Weidong},
  booktitle={Proceedings of the IEEE/CVF Conference on Computer Vision and Pattern Recognition},
  pages={8781--8791},
  year={2023}
}

@inproceedings{xu2024scorehypo,
  title={ScoreHypo: Probabilistic Human Mesh Estimation with Hypothesis Scoring},
  author={Xu, Yuan and Ma, Xiaoxuan and Su, Jiajun and Zhu, Wentao and Qiao, Yu and Wang, Yizhou},
  booktitle={Proceedings of the IEEE/CVF Conference on Computer Vision and Pattern Recognition},
  pages={979--989},
  year={2024}
}

@article{christiano2017rlhf,
  title={Deep reinforcement learning from human preferences},
  author={Christiano, Paul F and Leike, Jan and Brown, Tom and Martic, Miljan and Legg, Shane and Amodei, Dario},
  journal={Advances in neural information processing systems},
  volume={30},
  year={2017}
}

@article{rafailov2024dpo,
  title={Direct preference optimization: Your language model is secretly a reward model},
  author={Rafailov, Rafael and Sharma, Archit and Mitchell, Eric and Manning, Christopher D and Ermon, Stefano and Finn, Chelsea},
  journal={Advances in Neural Information Processing Systems},
  volume={36},
  year={2024}
}

@article{xu2024imagereward,
  title={Imagereward: Learning and evaluating human preferences for text-to-image generation},
  author={Xu, Jiazheng and Liu, Xiao and Wu, Yuchen and Tong, Yuxuan and Li, Qinkai and Ding, Ming and Tang, Jie and Dong, Yuxiao},
  journal={Advances in Neural Information Processing Systems},
  volume={36},
  year={2024}
}

@inproceedings{wallace2024diffusiondpo,
  title={Diffusion model alignment using direct preference optimization},
  author={Wallace, Bram and Dang, Meihua and Rafailov, Rafael and Zhou, Linqi and Lou, Aaron and Purushwalkam, Senthil and Ermon, Stefano and Xiong, Caiming and Joty, Shafiq and Naik, Nikhil},
  booktitle={Proceedings of the IEEE/CVF Conference on Computer Vision and Pattern Recognition},
  pages={8228--8238},
  year={2024}
}

@article{ho2020ddpmdiffusion,
  title={Denoising diffusion probabilistic models},
  author={Ho, Jonathan and Jain, Ajay and Abbeel, Pieter},
  journal={Advances in neural information processing systems},
  volume={33},
  pages={6840--6851},
  year={2020}
}

@article{song2020ddim,
  title={Denoising diffusion implicit models},
  author={Song, Jiaming and Meng, Chenlin and Ermon, Stefano},
  journal={arXiv preprint arXiv:2010.02502},
  year={2020}
}

@article{zhai2020iqasurvey,
  title={Perceptual image quality assessment: a survey},
  author={Zhai, Guangtao and Min, Xiongkuo},
  journal={Science China Information Sciences},
  volume={63},
  pages={1--52},
  year={2020},
  publisher={Springer}
}

@inproceedings{wang2023zolly,
  title={Zolly: Zoom focal length correctly for perspective-distorted human mesh reconstruction},
  author={Wang, Wenjia and Ge, Yongtao and Mei, Haiyi and Cai, Zhongang and Sun, Qingping and Wang, Yanjun and Shen, Chunhua and Yang, Lei and Komura, Taku},
  booktitle={Proceedings of the IEEE/CVF International Conference on Computer Vision},
  pages={3925--3935},
  year={2023}
}

@article{cai2024gtahuman,
  title={Playing for 3d human recovery},
  author={Cai, Zhongang and Zhang, Mingyuan and Ren, Jiawei and Wei, Chen and Ren, Daxuan and Lin, Zhengyu and Zhao, Haiyu and Yang, Lei and Loy, Chen Change and Liu, Ziwei},
  journal={IEEE Transactions on Pattern Analysis and Machine Intelligence},
  year={2024},
  publisher={IEEE}
}

@inproceedings{sengupta2023humaniflow,
  title={Humaniflow: Ancestor-conditioned normalising flows on so (3) manifolds for human pose and shape distribution estimation},
  author={Sengupta, Akash and Budvytis, Ignas and Cipolla, Roberto},
  booktitle={Proceedings of the IEEE/CVF Conference on Computer Vision and Pattern Recognition},
  pages={4779--4789},
  year={2023}
}

@inproceedings{foo2023hmdiff,
  title={Distribution-aligned diffusion for human mesh recovery},
  author={Foo, Lin Geng and Gong, Jia and Rahmani, Hossein and Liu, Jun},
  booktitle={Proceedings of the IEEE/CVF International Conference on Computer Vision},
  pages={9221--9232},
  year={2023}
}

@inproceedings{yin2023hi4d,
  title={Hi4d: 4d instance segmentation of close human interaction},
  author={Yin, Yifei and Guo, Chen and Kaufmann, Manuel and Zarate, Juan Jose and Song, Jie and Hilliges, Otmar},
  booktitle={Proceedings of the IEEE/CVF Conference on Computer Vision and Pattern Recognition},
  pages={17016--17027},
  year={2023}
}

@inproceedings{kocabas2021spec,
  title={SPEC: Seeing people in the wild with an estimated camera},
  author={Kocabas, Muhammed and Huang, Chun-Hao P and Tesch, Joachim and M{\"u}ller, Lea and Hilliges, Otmar and Black, Michael J},
  booktitle={Proceedings of the IEEE/CVF International Conference on Computer Vision},
  pages={11035--11045},
  year={2021}
}

@inproceedings{zheng2023potter,
  title={Potter: Pooling attention transformer for efficient human mesh recovery},
  author={Zheng, Ce and Liu, Xianpeng and Qi, Guo-Jun and Chen, Chen},
  booktitle={Proceedings of the IEEE/CVF Conference on Computer Vision and Pattern Recognition},
  pages={1611--1620},
  year={2023}
}

@article{biggs20203dmultibodies,
  title={3d multi-bodies: Fitting sets of plausible 3d human models to ambiguous image data},
  author={Biggs, Benjamin and Novotny, David and Ehrhardt, Sebastien and Joo, Hanbyul and Graham, Ben and Vedaldi, Andrea},
  journal={Advances in neural information processing systems},
  volume={33},
  pages={20496--20507},
  year={2020}
}

@inproceedings{sengupta2021hierarchical,
  title={Hierarchical kinematic probability distributions for 3D human shape and pose estimation from images in the wild},
  author={Sengupta, Akash and Budvytis, Ignas and Cipolla, Roberto},
  booktitle={Proceedings of the IEEE/CVF international conference on computer vision},
  pages={11219--11229},
  year={2021}
}

@inproceedings{lin2014mscoco,
  title={Microsoft coco: Common objects in context},
  author={Lin, Tsung-Yi and Maire, Michael and Belongie, Serge and Hays, James and Perona, Pietro and Ramanan, Deva and Doll{\'a}r, Piotr and Zitnick, C Lawrence},
  booktitle={Computer Vision--ECCV 2014: 13th European Conference, Zurich, Switzerland, September 6-12, 2014, Proceedings, Part V 13},
  pages={740--755},
  year={2014},
  organization={Springer}
}

@inproceedings{andriluka2014mpii,
  title={2d human pose estimation: New benchmark and state of the art analysis},
  author={Andriluka, Mykhaylo and Pishchulin, Leonid and Gehler, Peter and Schiele, Bernt},
  booktitle={Proceedings of the IEEE Conference on computer Vision and Pattern Recognition},
  pages={3686--3693},
  year={2014}
}

@inproceedings{mehta20173dhp,
  title={Monocular 3d human pose estimation in the wild using improved cnn supervision},
  author={Mehta, Dushyant and Rhodin, Helge and Casas, Dan and Fua, Pascal and Sotnychenko, Oleksandr and Xu, Weipeng and Theobalt, Christian},
  booktitle={2017 international conference on 3D vision (3DV)},
  pages={506--516},
  year={2017},
  organization={IEEE}
}

@inproceedings{kanazawa2019instavariety,
  title={Learning 3d human dynamics from video},
  author={Kanazawa, Angjoo and Zhang, Jason Y and Felsen, Panna and Malik, Jitendra},
  booktitle={Proceedings of the IEEE/CVF conference on computer vision and pattern recognition},
  pages={5614--5623},
  year={2019}
}

@inproceedings{cheng2023dnarender,
  title={Dna-rendering: A diverse neural actor repository for high-fidelity human-centric rendering},
  author={Cheng, Wei and Chen, Ruixiang and Fan, Siming and Yin, Wanqi and Chen, Keyu and Cai, Zhongang and Wang, Jingbo and Gao, Yang and Yu, Zhengming and Lin, Zhengyu and others},
  booktitle={Proceedings of the IEEE/CVF International Conference on Computer Vision},
  pages={19982--19993},
  year={2023}
}

@inproceedings{patel2025camerahmr,
  title={Camerahmr: Aligning people with perspective},
  author={Patel, Priyanka and Black, Michael J},
  booktitle={2025 International Conference on 3D Vision (3DV)},
  pages={1562--1571},
  year={2025},
  organization={IEEE}
}

@inproceedings{wang2025prompthmr,
  title={PromptHMR: Promptable Human Mesh Recovery},
  author={Wang, Yufu and Sun, Yu and Patel, Priyanka and Daniilidis, Kostas and Black, Michael J and Kocabas, Muhammed},
  booktitle={Proceedings of the Computer Vision and Pattern Recognition Conference},
  pages={1148--1159},
  year={2025}
}

@article{shen2025adhmr,
  title={ADHMR: Aligning Diffusion-based Human Mesh Recovery via Direct Preference Optimization},
  author={Shen, Wenhao and Yin, Wanqi and Yang, Xiaofeng and Chen, Cheng and Song, Chaoyue and Cai, Zhongang and Yang, Lei and Wang, Hao and Lin, Guosheng},
  journal={arXiv preprint arXiv:2505.10250},
  year={2025}
}

@inproceedings{fiche2025mega,
  title={MEGA: Masked Generative Autoencoder for Human Mesh Recovery},
  author={Fiche, Gu{\'e}nol{\'e} and Leglaive, Simon and Alameda-Pineda, Xavier and Moreno-Noguer, Francesc},
  booktitle={Proceedings of the Computer Vision and Pattern Recognition Conference},
  pages={5366--5378},
  year={2025}
}

@article{sarandi2024nlf,
  title={Neural localizer fields for continuous 3d human pose and shape estimation},
  author={S{\'a}r{\'a}ndi, Istv{\'a}n and Pons-Moll, Gerard},
  journal={Advances in Neural Information Processing Systems},
  volume={37},
  pages={140032--140065},
  year={2024}
}

@article{shao2024deepseekmath,
  title={Deepseekmath: Pushing the limits of mathematical reasoning in open language models},
  author={Shao, Zhihong and Wang, Peiyi and Zhu, Qihao and Xu, Runxin and Song, Junxiao and Bi, Xiao and Zhang, Haowei and Zhang, Mingchuan and Li, YK and others},
  journal={arXiv preprint arXiv:2402.03300},
  year={2024}
}

@article{lu2022dpm,
  title={Dpm-solver: A fast ode solver for diffusion probabilistic model sampling in around 10 steps},
  author={Lu, Cheng and Zhou, Yuhao and Bao, Fan and Chen, Jianfei and Li, Chongxuan and Zhu, Jun},
  journal={Advances in neural information processing systems},
  volume={35},
  pages={5775--5787},
  year={2022}
}

@article{xue2025dancegrpo,
  title={DanceGRPO: Unleashing GRPO on Visual Generation},
  author={Xue, Zeyue and Wu, Jie and Gao, Yu and Kong, Fangyuan and Zhu, Lingting and Chen, Mengzhao and Liu, Zhiheng and Liu, Wei and Guo, Qiushan and Huang, Weilin and others},
  journal={arXiv preprint arXiv:2505.07818},
  year={2025}
}

@article{guo2025deepseekr1,
  title={Deepseek-r1: Incentivizing reasoning capability in llms via reinforcement learning},
  author={Guo, Daya and Yang, Dejian and Zhang, Haowei and Song, Junxiao and Zhang, Ruoyu and Xu, Runxin and Zhu, Qihao and Ma, Shirong and Wang, Peiyi and Bi, Xiao and others},
  journal={arXiv preprint arXiv:2501.12948},
  year={2025}
}

@article{fan2023dpok,
  title={Dpok: Reinforcement learning for fine-tuning text-to-image diffusion models},
  author={Fan, Ying and Watkins, Olivia and Du, Yuqing and Liu, Hao and Ryu, Moonkyung and Boutilier, Craig and Abbeel, Pieter and Ghavamzadeh, Mohammad and Lee, Kangwook and Lee, Kimin},
  journal={Advances in Neural Information Processing Systems},
  volume={36},
  pages={79858--79885},
  year={2023}
}

@inproceedings{black2023ddpo,
  title={Training Diffusion Models with Reinforcement Learning},
  author={Black, Kevin and Janner, Michael and Du, Yilun and Kostrikov, Ilya and Levine, Sergey},
  booktitle={The Twelfth International Conference on Learning Representations},
  year={2024},
}

@inproceedings{clark2024draft,
  title={Directly Fine-Tuning Diffusion Models on Differentiable Rewards},
  author={Clark, Kevin and Vicol, Paul and Swersky, Kevin and Fleet, David J},
  booktitle={The Twelfth International Conference on Learning Representations},
  year={2024},
}

@inproceedings{yang2024d3po,
  title={Using human feedback to fine-tune diffusion models without any reward model},
  author={Yang, Kai and Tao, Jian and Lyu, Jiafei and Ge, Chunjiang and Chen, Jiaxin and Shen, Weihan and Zhu, Xiaolong and Li, Xiu},
  booktitle={Proceedings of the IEEE/CVF Conference on Computer Vision and Pattern Recognition},
  pages={8941--8951},
  year={2024}
}

@inproceedings{liang2025spo,
  title={Aesthetic post-training diffusion models from generic preferences with step-by-step preference optimization},
  author={Liang, Zhanhao and Yuan, Yuhui and Gu, Shuyang and Chen, Bohan and Hang, Tiankai and Cheng, Mingxi and Li, Ji and Zheng, Liang},
  booktitle={Proceedings of the Computer Vision and Pattern Recognition Conference},
  pages={13199--13208},
  year={2025}
}

@article{li2024diffusionkto,
  title={Aligning diffusion models by optimizing human utility},
  author={Li, Shufan and Kallidromitis, Konstantinos and Gokul, Akash and Kato, Yusuke and Kozuka, Kazuki},
  journal={Advances in Neural Information Processing Systems},
  volume={37},
  pages={24897--24925},
  year={2024}
}

@article{dubois2023alpacafarm,
  title={Alpacafarm: A simulation framework for methods that learn from human feedback},
  author={Dubois, Yann and Li, Chen Xuechen and Taori, Rohan and Zhang, Tianyi and Gulrajani, Ishaan and Ba, Jimmy and Guestrin, Carlos and Liang, Percy S and Hashimoto, Tatsunori B},
  journal={Advances in Neural Information Processing Systems},
  volume={36},
  pages={30039--30069},
  year={2023}
}

@article{li2024crowdsourced,
  title={From crowdsourced data to high-quality benchmarks: Arena-hard and benchbuilder pipeline},
  author={Li, Tianle and Chiang, Wei-Lin and Frick, Evan and Dunlap, Lisa and Wu, Tianhao and Zhu, Banghua and Gonzalez, Joseph E and Stoica, Ion},
  journal={arXiv preprint arXiv:2406.11939},
  year={2024}
}

@inproceedings{qin2024large,
  title={Large language models are effective text rankers with pairwise ranking prompting},
  author={Qin, Zhen and Jagerman, Rolf and Hui, Kai and Zhuang, Honglei and Wu, Junru and Yan, Le and Shen, Jiaming and Liu, Tianqi and Liu, Jialu and Metzler, Donald and others},
  booktitle={Findings of the Association for Computational Linguistics: NAACL 2024},
  pages={1504--1518},
  year={2024}
}

@article{verga2024replacing,
  title={Replacing judges with juries: Evaluating llm generations with a panel of diverse models},
  author={Verga, Pat and Hofstatter, Sebastian and Althammer, Sophia and Su, Yixuan and Piktus, Aleksandra and Arkhangorodsky, Arkady and Xu, Minjie and White, Naomi and Lewis, Patrick},
  journal={arXiv preprint arXiv:2404.18796},
  year={2024}
}

@article{zhang2025sentient,
  title={Sentient Agent as a Judge: Evaluating Higher-Order Social Cognition in Large Language Models},
  author={Zhang, Bang and Ma, Ruotian and Jiang, Qingxuan and Wang, Peisong and Chen, Jiaqi and Xie, Zheng and Chen, Xingyu and Wang, Yue and Ye, Fanghua and Li, Jian and others},
  journal={arXiv preprint arXiv:2505.02847},
  year={2025}
}

@article{chan2023chateval,
  title={Chateval: Towards better llm-based evaluators through multi-agent debate},
  author={Chan, Chi-Min and Chen, Weize and Su, Yusheng and Yu, Jianxuan and Xue, Wei and Zhang, Shanghang and Fu, Jie and Liu, Zhiyuan},
  journal={arXiv preprint arXiv:2308.07201},
  year={2023}
}

@inproceedings{huang2025empirical,
  title={An empirical study of llm-as-a-judge for llm evaluation: Fine-tuned judge model is not a general substitute for gpt-4},
  author={Huang, Hui and Bu, Xingyuan and Zhou, Hongli and Qu, Yingqi and Liu, Jing and Yang, Muyun and Xu, Bing and Zhao, Tiejun},
  booktitle={Findings of the Association for Computational Linguistics: ACL 2025},
  pages={5880--5895},
  year={2025}
}

@article{zhu2023judgelm,
  title={Judgelm: Fine-tuned large language models are scalable judges},
  author={Zhu, Lianghui and Wang, Xinggang and Wang, Xinlong},
  journal={arXiv preprint arXiv:2310.17631},
  year={2023}
}

@inproceedings{kim2023prometheus,
  title={Prometheus: Inducing fine-grained evaluation capability in language models},
  author={Kim, Seungone and Shin, Jamin and Cho, Yejin and Jang, Joel and Longpre, Shayne and Lee, Hwaran and Yun, Sangdoo and Shin, Seongjin and Kim, Sungdong and Thorne, James and others},
  booktitle={The Twelfth International Conference on Learning Representations},
  year={2023}
}

@article{wang2025trustjudge,
  title={TrustJudge: Inconsistencies of LLM-as-a-Judge and How to Alleviate Them},
  author={Wang, Yidong and Song, Yunze and Zhu, Tingyuan and Zhang, Xuanwang and Yu, Zhuohao and Chen, Hao and Song, Chiyu and Wang, Qiufeng and Wang, Cunxiang and Wu, Zhen and others},
  journal={arXiv preprint arXiv:2509.21117},
  year={2025}
}

@article{liu2025flowgrpo,
  title={Flow-grpo: Training flow matching models via online rl},
  author={Liu, Jie and Liu, Gongye and Liang, Jiajun and Li, Yangguang and Liu, Jiaheng and Wang, Xintao and Wan, Pengfei and Zhang, Di and Ouyang, Wanli},
  journal={arXiv preprint arXiv:2505.05470},
  year={2025}
}

@inproceedings{song2021score,
  title={Score-Based Generative Modeling through Stochastic Differential Equations},
  author={Song, Yang and Sohl-Dickstein, Jascha and Kingma, Diederik P and Kumar, Abhishek and Ermon, Stefano and Poole, Ben},
  booktitle={International Conference on Learning Representations},
  year={2021}
}

@article{auer2002ucb1,
  title={Finite-time analysis of the multiarmed bandit problem},
  author={Auer, Peter and Cesa-Bianchi, Nicolo and Fischer, Paul},
  journal={Machine learning},
  volume={47},
  number={2},
  pages={235--256},
  year={2002},
  publisher={Springer}
}

@inproceedings{shankar2024validates,
  title={Who validates the validators? aligning llm-assisted evaluation of llm outputs with human preferences},
  author={Shankar, Shreya and Zamfirescu-Pereira, JD and Hartmann, Bj{\"o}rn and Parameswaran, Aditya and Arawjo, Ian},
  booktitle={Proceedings of the 37th Annual ACM Symposium on User Interface Software and Technology},
  pages={1--14},
  year={2024}
}

@inproceedings{li2025generation,
  title={From generation to judgment: Opportunities and challenges of llm-as-a-judge},
  author={Li, Dawei and Jiang, Bohan and Huang, Liangjie and Beigi, Alimohammad and Zhao, Chengshuai and Tan, Zhen and Bhattacharjee, Amrita and Jiang, Yuxuan and Chen, Canyu and Wu, Tianhao and others},
  booktitle={Proceedings of the 2025 Conference on Empirical Methods in Natural Language Processing},
  pages={2757--2791},
  year={2025}
}
}


\end{document}